\documentclass{article} % For LaTeX2e
\PassOptionsToPackage{table}{xcolor}
\usepackage{iclr2025_conference,times}

\usepackage{amsmath,amsfonts,bm}

\def\eqref#1{equation~\ref{#1}}
\def\1{\bm{1}}

\DeclareMathAlphabet{\mathsfit}{\encodingdefault}{\sfdefault}{m}{sl}
\SetMathAlphabet{\mathsfit}{bold}{\encodingdefault}{\sfdefault}{bx}{n}

\usepackage{hyperref}
\usepackage{url}
\usepackage{graphicx}
\usepackage{booktabs} 
\usepackage{enumitem}
\usepackage{pifont}
\usepackage{multirow}
\usepackage{subcaption}
\usepackage{wrapfig}
\usepackage{bbm}
\usepackage[table]{xcolor}
\usepackage{xspace}
\usepackage[most]{tcolorbox}

\NewDocumentCommand{\zefan}
{ mO{} }{\textcolor{green}{\textsuperscript{\textit{zefan}}\textsf{\textbf{\small[#1]}}}}
\NewDocumentCommand{\lifan}
{ mO{} }{\textcolor{cyan}{\textsuperscript{\textit{lifan}}\textsf{\textbf{\small[#1]}}}}

\newcommand{\methodname}{\textbf{RLPR}\xspace}

\newcommand{\github}{\raisebox{-1.5pt}{\includegraphics[height=1.05em]{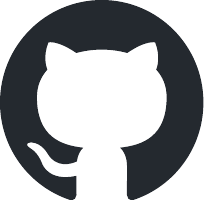}}\xspace}
\newcommand{\huggingface}{\raisebox{-1.5pt}{\includegraphics[height=1.05em]{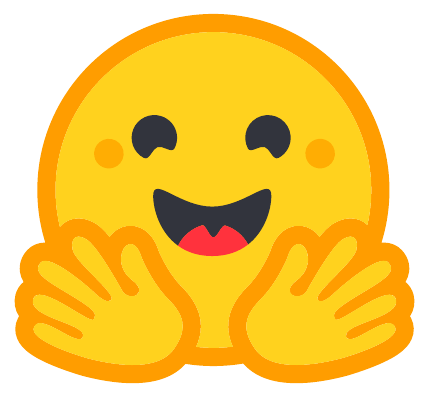}}\xspace}

\title{RLPR: Extrapolating RLVR to general domains without verifiers}

\author{
Tianyu Yu\,$^{1\dagger}$\thanks{Project Lead.} \quad Bo Ji\,$^{2}$\thanks{Core Contributors.} \quad Shouli Wang\,$^{4\dagger}$ \quad Shu Yao\,$^{5\dagger}$ \quad Zefan Wang\,$^{1\dagger}$ \\
\textbf{ Ganqu Cui}\,$^{1}$ \quad \textbf{Lifan Yuan}\,$^{6}$ \quad \textbf{Ning Ding}\,$^{1}$ \quad 
\textbf{Yuan Yao}\,$^{2,3}$\thanks{Corresponding authors.} \quad \textbf{Zhiyuan Liu}\,$^{1\ddagger}$ \\
\textbf{ Maosong Sun}\,$^{1}$ \quad \textbf{Tat-Seng Chua}\,$^{2}$\\
$^1$Tsinghua University \quad $^2$National University of Singapore \quad $^3$Shanghai Qi Zhi Institute \\
$^4$Harbin Institute of Technology \quad $^5$Beijing University of Posts and Telecommunications \\
$^6$University of Illinois Urbana-Champaign \\
\vspace{2mm}
\texttt{yiranytianyu@gmail.com \quad yaoyuanthu@gmail.com}\\
\vspace{1mm}
\github \href{https://github.com/openbmb/RLPR}{{\text{RLPR Code}}} 
\hspace{25mm} \huggingface \href{https://huggingface.co/datasets/openbmb/RLPR-Train-Dataset}{{\text{RLPR Dataset}}} 
\hspace{25mm} \huggingface \href{https://huggingface.co/collections/openbmb/rlpr-6857fa5d22cbe64327a3f8f6}{{\text{RLPR Models}}} 
}

\definecolor{myred}{rgb}{1.0, 0.25, 0}
\newcommand{\deltavalue}[3]{\hspace{#3mm}#1\scalebox{0.7}{\textcolor[HTML]{a6192e}{{-#2}}}}

\iclrfinalcopy % Uncomment for camera-ready version, but NOT for submission.

\usepackage{fancyhdr}
\fancypagestyle{myfancy}{
  \fancyhf{}  
  \fancyhead[L]{RLPR: Extrapolating RLVR to General Domains without Verifiers}
  \fancyfoot[C]{\thepage}
}
\begin{document}

\maketitle

\thispagestyle{myfancy}
\pagestyle{myfancy}

\begin{figure}[h]
    \centering
    \vspace*{-5mm}
    \includegraphics[width=\textwidth]{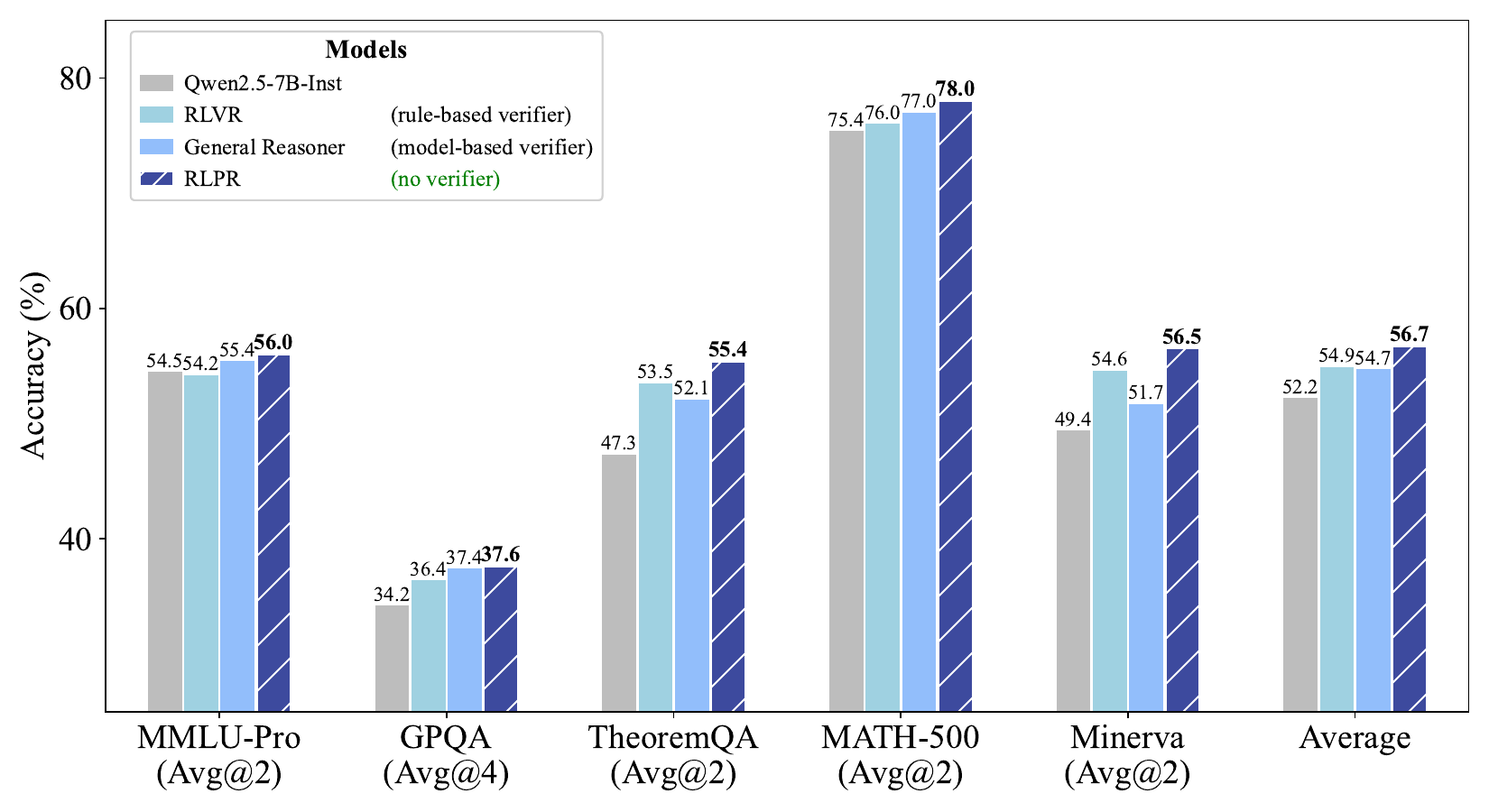}
    \vspace*{-5mm}
    \label{fig:performance}
    \caption{Overall performance on general-domain and mathematical reasoning benchmarks. By simply replacing the rule-based verifier reward of RLVR with the proposed LLM's intrinsic probability reward, \methodname~achieves consistent improvements in both mathematical and general domains, even outperforming strong RL methods driven by model-based verifier reward. Average: average accuracy of five benchmarks. Verifier requirements of different methods are listed in parentheses.}
    \label{fig:performance_viper}
\end{figure}

\begin{abstract}

Reinforcement Learning with Verifiable Rewards (RLVR) demonstrates promising potential in advancing the reasoning capabilities of LLMs. However, its success remains largely confined to mathematical and code domains. This primary limitation stems from the heavy reliance on domain-specific verifiers, which results in prohibitive complexity and limited scalability. To address the challenge, our key observation is that LLM's intrinsic probability of generating a correct free-form answer directly indicates its own evaluation of the reasoning reward (i.e., how well the reasoning process leads to the correct answer). Building on this insight, we propose \methodname, a simple verifier-free framework that extrapolates RLVR to broader general domains. \methodname~uses the LLM's own token probability scores for reference answers as the reward signal and maximizes the expected reward during training. We find that addressing the high variance of this noisy probability reward is crucial to make it work, and propose prob-to-reward and stabilizing methods to ensure a precise and stable reward from LLM intrinsic probabilities. Comprehensive experiments in four general-domain benchmarks and three mathematical benchmarks show that \methodname~consistently improves reasoning capabilities in both areas for Gemma, Llama, and Qwen based models. Notably, \methodname~outperforms concurrent VeriFree by 7.6 points on TheoremQA and 7.5 points on Minerva, and even surpasses strong verifier-model-dependent approaches General-Reasoner by 1.6 average points across seven benchmarks.
\end{abstract}

\section{Introduction}

Large-scale Reinforcement Learning with Verifiable Rewards~(RLVR) has emerged as a promising paradigm to advance the reasoning capabilities of Large Language Models (LLMs)~\citep{openai2024openaio1card,deepseekai2025deepseekr1incentivizingreasoningcapability,hu2025openreasonerzeroopensourceapproach,deepcoder2025}. This paradigm not only shows the power of scaling test-time computation for addressing complex problems, but also sheds valuable light on paths to AGI with incentivized exploration and evolution.

\begin{figure}[t]
    \centering
    \includegraphics[width=\linewidth]{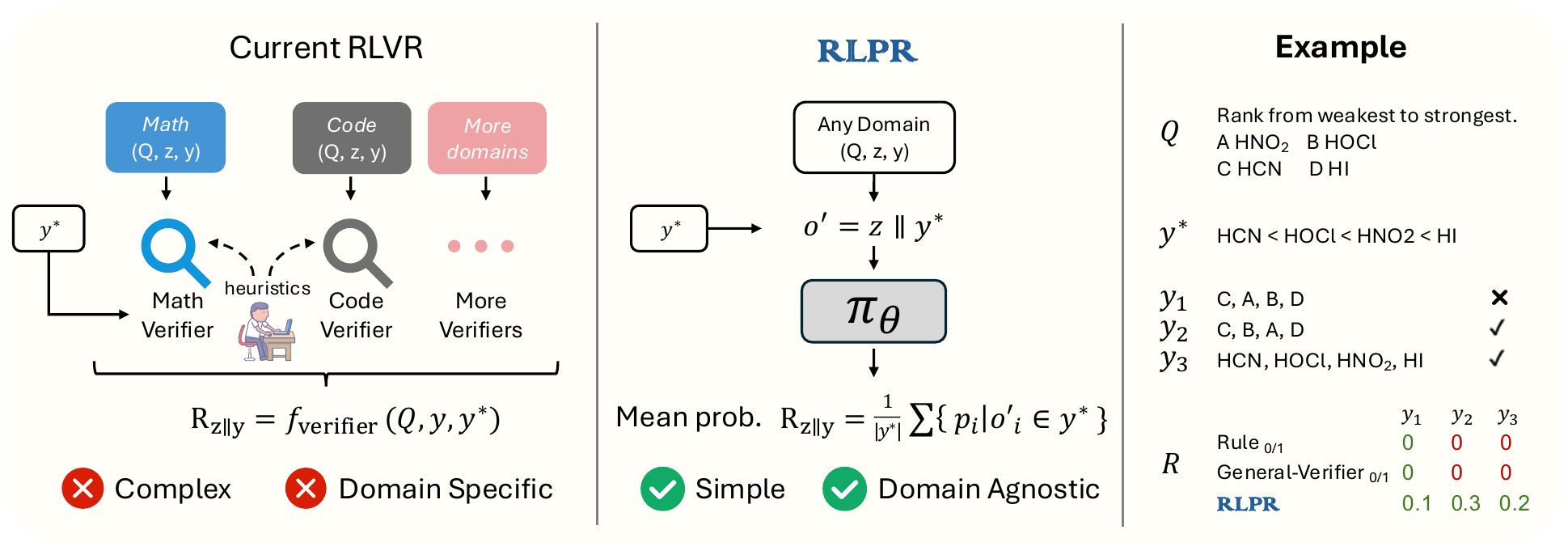}
    \caption{Existing RLVR methods rely on specialized verifiers for each domain, suffering from high complexity and limited scalability. We propose the \methodname~framework, which replaces the complex verifier-based reward with a simple probability-based reward generated by the policy model $\pi_\theta$. $Q$: input question, $z$: generated reasoning content before final answer, $y$: generated final answer, $y^*$: reference answer. As shown in the example on the right side, rules and verifier models wrongly label both $y_2$ and $y_3$ as incorrect due to their limited capability of handling natural language complexity.}
    \label{fig:teaser}
\end{figure}

However, in contrast to the pretraining of LLMs that can learn foundational capabilities from general domain data, most RLVR methods are confined to mathematics~\citep{hu2025openreasonerzeroopensourceapproach,liu2025understandingr1zeroliketrainingcritical,zeng2025simplerlzooinvestigatingtamingzero,yu2025dapo} and code generation~\citep{deepcoder2025,skywork-or1-2025,cui2025process}. The primary reason is that existing RLVR methods heavily rely on domain-specific verifiers to obtain reward, as shown in Figure~\ref{fig:teaser}. The most widely adopted verifiers are handcrafted rules~\citep{hu2025openreasonerzeroopensourceapproach,liu2025understandingr1zeroliketrainingcritical,zeng2025simplerlzooinvestigatingtamingzero}. Extending these rule-based reward systems to new models and domains typically requires prohibitive heuristic engineering. Moreover, for general-domain reasoning with free-form answers, it is even impossible to devise rule-based verifiers due to the high diversity and complexity of natural language. Recent works attempt to address this problem by training specialized LLMs as verifier models~\citep{ma2025generalreasoneradvancingllmreasoning}. However, training LLMs for general reward evaluation requires non-trivial and extensive data annotation, which often leads to unsatisfactory reward quality in practice. Involving separate verifier models also complicates the RL training framework and introduces additional computation cost. As a result, this scalability problem prevents existing RLVR methods from utilizing rich general-domain data and limits the potential of broader reasoning capabilities.

To address the problem, we propose the \methodname~framework~(\textbf{R}einforcement \textbf{L}earning with Reference \textbf{P}robability \textbf{R}eward) that extrapolates general-domain RLVR without external verifiers. \textit{The basic insight is that LLM's intrinsic probability of generating a correct answer directly indicates its own evaluation of the reasoning reward} (i.e., how well the reasoning process leads to the correct answer). It also reflects the policy by measuring how likely the LLM is to take the correct action. Therefore, we can directly leverage this probability signal as a reward to incentivize reasoning for the correct answer in general domains. Since this probability score is a natural built-in of LLM's foundational capabilities, it offers good coverage and potential for reward evaluation even without any specialized fine-tuning. It can also better deal with the complexity and diversity of free-form natural language answers, giving reasonable reward even to partially correct answers.

Specifically, \methodname~introduces two key innovations: (1) At the reward modeling level, we propose a simple and scalable alternative to the explicit reward from external verifiers with an intrinsic Probability-based Reward~(PR), calculated by the average decoding probabilities of the reference answer tokens.
Compared with naive sequence likelihood as reward~\citep{zhou2025reinforcing}, the proposed PR shows better robustness and higher reward quality on quantitative examinations~(see Figure~\ref{fig:pr_auc}).
Moreover, we propose a simple debiasing method to eliminate the reward bias from text by optimizing the reward advantage over the same prompt without reasoning.
(2) At the training level, we propose an adaptive curriculum learning mechanism to stabilize training. We adaptively remove prompts yielding low reward standard deviation (indicating prompts that are too easy or too complex), using a dynamic threshold based on the exponential moving average of past rewards' standard deviation. We find that this approach can well keep up with the reward distribution shifts during training, and improves both the training stability and final performance.

Comprehensive experiments on seven benchmarks show that, without any external verifiers, \methodname~substantially enhances reasoning capabilities in both mathematical and general domains. 
Leveraging Qwen2.5-7B~\citep{qwen2.5} as base model, \methodname~achieves 56.0 on MMLU-Pro and 55.4 on TheoremQA, even surpassing the strong General Reasoner-7B~\citep{ma2025generalreasoneradvancingllmreasoning} that utilizes a specially trained 1.5B verifier model. 
Furthermore, compared with VeriFree~\citep{zhou2025reinforcing}, a concurrent verifier-free approach, \methodname~achieves significant improvement of 7.6 on TheoremQA and 7.5 on Minerva. We also evaluate \methodname~on models from Llama3.1~\cite{grattafiori2024llama3herdmodels} and Gemma2~\cite{gemmateam2024gemma2improvingopen}, achieving improvements of 6.4 and 6.1 average points across seven benchmarks respectively.

The contribution of this work can be summarized as fourfold: 
(1) We present \methodname, a simple and scalable framework that extends RLVR to general domains without using external verifiers. 
(2) We propose a novel probability reward that eliminates the need for external verifiers and achieves better reward quality than naive likelihood as a reward. 
(3) We introduce a novel standard deviation filtering strategy that effectively stabilizes training by removing samples with low reward standard deviation. 
(4) We conduct comprehensive experiments to demonstrate the effectiveness of the proposed framework on various base models from Qwen, Llama and Gemma. 
All the codes, data, and model weights are released to facilitate future research.

\section{\methodname}

In this section, we first introduce the fundamentals of RLVR and describe the procedure to calculate the probability reward for \methodname. Then we introduce the debiasing method and the standard deviation filtering approach.

\subsection{Reinforcement Learning from Verifiable Rewards}
Reinforcement learning from verifiable reward~(RLVR) is a general post-training paradigm in which a rule-based verifier assigns a scalar reward score to each generated response. 
Specifically, given a prompt \({x}\), the policy \(\pi_\theta\) produces reasoning content \({z}\) and the final answer \({y}\). 
Then the expected verifier score is optimized: 
\begin{equation}
\begin{aligned}
\mathcal{J}(\theta) &= \mathbb{E}_{{z},{y}\sim \pi_{\theta}{(\cdot|{x})}}\left[f_{\text{verifier}}({y},{y}^*)\right],
\label{obj}
\end{aligned}
\end{equation}
where $f_{\text{verifier}}$ is a task-specific, rule-based verifier checking whether the generated answer $y$ passes the test defined by ground truth $y^*$. 
Common instantiations include symbolic verifiers~\citep{mathverify} for mathematical problems or sandboxed execution~\citep{bytedanceseedfoundationcodeteam2025fullstackbenchevaluatingllms} for code generation. 
However, building rule-based verifiers is a laborious, systematic effort that involves designing handcrafted rules and edge case handling. 
This restricts the application of RLVR to new domains.

\subsection{Probability Reward}

Motivated by the observation that the LLM's intrinsic probability of generating a correct answer directly indicates its internal evaluation of the reasoning quality, we use per-token decoding probabilities of the reference answer as the reward signal. Unlike existing methods that rely on domain-specific verifiers~\citep{cui2025process, deepcoder2025}, which require substantial manual heuristics and engineering effort for the construction of verifiers, our reward computation process involves only the model itself.
An overview of the process is illustrated in Figure~\ref{fig:teaser}.

We denote each response to question $Q$ with $o=(o_0, \cdots,o_N)$, where $o_i$ is an individual token in the response. 
To obtain probabilities, we first extract the generated answer $y$ from the full response sequence and denote the remaining content as reasoning $z$. We then construct a modified sequence $o'=(o'_0, \cdots, o'_{N'})$ by replacing the generated answer with the reference from the training data. This sequence is fed to the policy model to get probabilities $(p_0, \cdots, p_{N'})$. The probability reward is computed as:

\begin{equation}
\label{eq:prob_score}
    r =  f_\text{seq}(\{p_i | o'_i\in y^*\}),
\end{equation}
where $f_\text{seq}$ aggregates per-token probabilities into a single reward scalar for the response $o$. 
While using $f_\text{seq}=\sqrt[n]{\prod}$ (the normalized product of probabilities, i.e., sequence likelihood) reflects the overall likelihood of the reference answer, we observe that it introduces high variance and is overly sensitive to minor variations, such as synonyms. 
For instance, the token probability sequences (0.01, 0.7, 0.9) and (0.05, 0.7, 0.9) yield vastly different scores under the product, despite only a small difference on the first token.  To address this issue, we instead adopt $f_\text{seq}=\frac{1}{\lvert~y^*\rvert}\sum$ (mean  probabilities), which yields a more robust reward signal and demonstrates superior correlation with answer quality in our analyses (see Fig~\ref{fig:pr_auc}). 
We observe that probability reward values are highly consistent with the quality of generated answer $y$, where high rewards are gained when the predicted answer is semantically similar to the reference answer and low rewards are assigned otherwise. Note that the length-normalization step is redundant for GRPO~\citep{shao2024deepseekmathpushinglimitsmathematical} but could be crucial for algorithms like REINFORCE++~\citep{hu2025reinforceefficientrlhfalgorithm} which do not include group-normalization.

\subsection{Reward Debiasing}

Although the probability-based rewards correlate strongly with response quality, they are also influenced by various latent factors.
We denote the contributors to the probability reward $r$ as $U_{r}$, which can be essentially decomposed into two latent factors:

\begin{equation}
    U_{r} =  U_{z} + U_{\text{others}},
\end{equation}
where $U_{z}$ represents the effects of the reasoning content, and $U_{\text{others}}$ captures the characteristics of other related factors, including the question and reference answer.
Using $r$ directly as a reward introduces bias associated with the unobserved factor $U_\text{other}$, potentially degrading the reward quality. To mitigate this, we introduce a base score $r'$ by computing the probability score of directly decoding the reference answer $y^*$, without intermediate reasoning $z$, using Eq~\ref{eq:prob_score}.  This gives $U_{z} = U_{r} - U_{r'}$, and the debiased probability reward is then calculated as with:

\begin{equation}
    \hat{r} = \operatorname{clip}(0, 1, r - r'),
\end{equation}
where the clipping operation ensures that the reward remains within a favorable numeric range $[0, 1]$.
This formulation effectively removes the potential bias from $U_{Q}$ and $U_{y^*}$ and models PR as the improvement in probability given the generated reasoning $z$. 
We observe this debiasing step stabilizes training and enhances reward robustness. 
The final gradient estimator of our objective function is: 
\begin{align}
\nabla\mathcal{J}_\text{RLPR}(\theta) &= \nabla\mathbb{E}_{o\sim\pi_{\theta}{(\cdot|{x})}}\left[\hat{r}\right] \nonumber
\\
&=\sum_{o}\hat{r}\ \pi_{\theta}(o|x) \nabla \log \pi_{\theta}(o|x) \nonumber\\
&=\mathbb{E}_{{o}\sim\pi_{\theta}(\cdot|{x})} \left [\hat{r}  \nabla\log\pi_\theta(o|x) \right]
\label{final_obj},
\end{align}
where we optimize the expected reward on the whole response $o=z||y$. 

\subsection{Standard deviation filtering}

Existing RLVR methods employ accuracy filtering~\citep{cui2025process} to stabilize training by excluding too difficult and too easy prompts. Typically, this involves filtering entirely correct or incorrect prompts.
However, the continuous nature of PR makes it challenging to directly apply accuracy filtering since it is hard to set a universal threshold for response correctness.

Through the analysis of accuracy filtering, we observe that filtering prompts with low standard deviation in reward values can effectively achieve a similar effect. Specifically, prompts that consistently yield all high or all low scores exhibit low standard deviation due to the boundedness of PR (i.e., all reward values lie within $[0, 1]$). 
Meanwhile, the overall standard deviation distribution continuously shifts during training, and a fixed threshold may cause either too strict or loose filtering at different training stages. To address this, we adopt an exponential moving average to dynamically update the filtering threshold $\beta$ using the average standard deviation of each training step. By filtering the prompts whose reward standard deviation is less than $\beta$, we introduce an adaptive curriculum learning mechanism to improve both the training stability and final performance.

\section{Experiments}

In this section, we empirically investigate the effectiveness of \methodname~in enhancing LLM reasoning capabilities. In addition to evaluating model performance, we also analyze reward quality of our proposed PR, the efficacy of different components, and the potential of applying \methodname~to verifiable domains such as mathematics.

\subsection{Experimental Setup}\label{sec:exp_setup}

\textbf{Models.} We conduct experiments on Gemma2~\cite{gemmateam2024gemma2improvingopen}, Llama3.1~\cite{grattafiori2024llama3herdmodels} and Qwen2.5~\citep{qwen2.5} series models for fair comparison with most existing methods and thorough evaluation. Unless otherwise specified, experiments are conducted on Qwen2.5-7B-Base.

\textbf{Training Data.} We adopt the collection of prompts released by \citep{ma2025generalreasoneradvancingllmreasoning}, which includes high-quality reasoning questions across multiple complex domains. To focus on the effectiveness of \methodname~in general domains, we only use non-mathematics prompts from the data. We ask GPT-4.1~\citep{openai2025gpt41} to filter out prompts that are too easy and finally get 77k prompts for training.

\textbf{Evaluation.} We evaluate the reasoning capabilities with multiple general reasoning and mathematical benchmarks. For math reasoning, we include MATH-500~\citep{cobbe2021math}, Minerva~\citep{lewkowycz2022minerva}, and AIME24. For general domains, we adopt four benchmarks: 

\begin{itemize}[leftmargin=2em]
\item  MMLU-Pro~\citep{wang2024mmluprorobustchallengingmultitask} is a widely used multitask language understanding benchmark that includes challenging, reasoning-intensive questions across diverse domains. We randomly sample 1000 prompts from the benchmark to strike a balance between efficiency and variance.

\item  GPQA~\citep{rein2023gpqagraduatelevelgoogleproofqa} includes graduate-level questions from multiple disciplines, including physics, chemistry, etc. We use the highest-quality GPQA-diamond subset for evaluation.

\item TheoremQA~\citep{chen2023theoremqatheoremdrivenquestionanswering} assesses a model’s ability to apply theorems to solve complex science problems. This benchmark includes 800 high-quality questions covering 350 theorems from domains including Math, Physics, etc. We remove the 53 multimodal instructions.

\item  WebInstruct. We hold out a validation split from WebInstruct~\citep{ma2025generalreasoneradvancingllmreasoning} as a more accessible benchmark for medium-sized models. Unlike the aforementioned benchmarks, this benchmark is designed to be less challenging while still assessing multidisciplinary reasoning.  We uniformly sample 1k prompts from the training set and remove potential data contamination by applying 10-gram deduplication, resulting in 638 distinct questions. 

\end{itemize}

\textbf{Baselines.} We compare our approach with the following established and contemporaneous methods: (1) Base models and Instruct models. We include the Qwen2.5~\citep{qwen2.5} model family for comparison, reporting results for both Qwen2.5-7B and Qwen2.5-7B-Instruct. We also compare with Gemma2-2B-it and Llama3.1-8B-Inst.
(2) PRIME~\citep{cui2025process} enhances the mathematical and code reasoning capabilities using implicit rewards.
(3) SimpleRL-Zoo~\citep{zeng2025simplerlzooinvestigatingtamingzero} trains the model using rule-based rewards. We report both results of the Qwen2.5-Math and Qwen2.5-7B as the base model.
(4) Oat-Zero~\citep{liu2025understandingr1zeroliketrainingcritical} proposes to remove the standard deviation and token-level normalization in GRPO.
(5) TTRL~\citep{zuo2025ttrltesttimereinforcementlearning} eliminates the reliance on labeled reference answers and instead uses majority voting to assign pseudo-labels to sampled responses. We report the result of the model trained on MATH-500~\citep{zuo2025ttrltesttimereinforcementlearning} prompts.
(6) General Reasoner~\citep{ma2025generalreasoneradvancingllmreasoning} conducts RLVR in multiple domains by introducing an additional verifier model, which is distilled from Gemini 2.0~\citep{gemini2024} to verify general-domain responses. 
(7) VeriFree~\citep{zhou2025reinforcing} is a concurrent work that uses the likelihood of reference answers (for those shorter than 7-tokens) as the reward signal and incorporates an auxiliary fine-tuning loss. As results were only released for the Qwen3~\citep{qwen3technicalreport} model series, we reproduce their method on Qwen2.5-7B using the official repository. For fair comparison, we evaluate both their provided prompt and our training prompt, finding that the original prompt yields better results. Therefore, we adopt this configuration for this baseline.

\textbf{Implementation Details.} We adopt the verl~\citep{sheng2024hybridflow} framework for efficient training. In each rollout step, we sample eight responses per prompt for a batch of 768 prompts using a temperature of 1, and subsequently perform 4 policy updates on the collected responses. The scale $\beta$ used for filtering is set to 0.5. The clip threshold in PPO loss is set to (0.8, 1.27) to prevent entropy collapse~\citep{yu2025dapo,cui2025entropy}. During evaluation, we set the rollout temperature to 1. To reduce the evaluation variance, we evaluate the model on each benchmark multiple times and report the final Avg@k results. The max generation length for training and evaluation is 3072, with minimal truncation observed. For baseline evaluation, we adopt the default generation temperature from the original papers. For baseline evaluation, we follow the corresponding papers to select generation parameters and use our setup if the original paper uses greedy decoding. 
For reliable answer extraction, we adopt the ``\textless{}think\textgreater{}\textless{}/think\textgreater{}\textless{}answer\textgreater{}\textless{}/answer\textgreater{}'' template of R1~\citep{liu2025understandingr1zeroliketrainingcritical} during training and use the striped content inside answer tags as the generated answer. For experiments on Gemma and Llama, we change the training and evaluation temperature to 0.6 and remove the \textless{}think\textgreater{} part in templates to prevent generation degradation. We observe that rule-based scoring scripts introduce errors in benchmarks containing question formats beyond multiple-choice. To address this, we deploy a Qwen2.5-7B-Inst model server for evaluation, and additionally leverage GPT-4.1 for more complex benchmarks, such as TheoremQA and Minerva.

\subsection{Main Results}

\begin{table}
    \centering
    \resizebox{\linewidth}{!}{
    \setlength\tabcolsep{2.2pt}
    \begin{tabular}{lcc|ccccccc|ccc}
    \toprule
     \multirow{2}{*}{\textbf{Model}} & \multirow{2}{*}{\textbf{Base}} & \multirow{2}{*}{\textbf{Verifier}}   & \textbf{MMLU-Pro} & \textbf{GPQA} & \textbf{TheoremQA} & \textbf{WebInst.} & \textbf{MATH-500} & \textbf{Minerva} & \textbf{AIME 24} & \textbf{General} & \textbf{All}\\
 & & & Avg@2 & Avg@4 & Avg@2 & Avg@2 & Avg@2 & Avg@2 & Avg@16 & - & -\\
 \midrule
  \multicolumn{12}{c}{\textbf{Gemma Models}} \\
 \midrule
 Gemma2-2B-it & Base & -- & 27.9 & 19.3 & 16.4 & 33.5 & 26.6 & 15.9 & 0.0 & 24.3 & 19.9 \\
 RLVR & Inst & Rule & 31.6 & 25.8 & 20.1 & \textbf{52.3} & \textbf{30.7} & 16.5 & \textbf{0.2} & 32.4 & 25.3  \\
\rowcolor[HTML]{D7E8E8}
 \methodname~& Inst & \textcolor{teal}{\ding{55}} & \textbf{33.5} & \textbf{28.5} & \textbf{21.2} & 52.0 & 30.4 & \textbf{17.1} & \textbf{0.2} & \textbf{33.8} & \textbf{26.0}\\ 

 \midrule
 \multicolumn{12}{c}{\textbf{Llama Models}} \\
 \midrule
 Llama3.1-8B-Inst & Base & -- & 46.4 & 31.6 & 31.3 & 54.7 & 50.1 & 32.7 & 4.2 & 40.5 & 35.6\\
 RLVR & Inst & Rule & 49.3 & 36.0 & 32.0 & 60.2 & 51.9 & 35.2 & 4.6 & 44.4 & 38.5 \\
\rowcolor[HTML]{D7E8E8}
 \methodname~& Inst & \textcolor{teal}{\ding{55}} &  \textbf{53.6} & \textbf{36.5} &  \textbf{35.5} & \textbf{68.5} & \textbf{54.1} & \textbf{39.0} & \textbf{8.8} & \textbf{48.5} & \textbf{42.3}\\
\midrule
 \multicolumn{12}{c}{\textbf{Qwen Models}} \\
 \midrule
Qwen2.5-7B & -- & -- & 45.3 & 32.4 & 41.4 & 60.4 & 63.0 & 37.6 & \hspace{2mm}6.5 & 44.9 & 40.9 \\
 Qwen2.5-7B-Inst & Base & -- & 54.5 & 34.2 & 47.3 & 72.6 & 75.4  & 49.4 & \hspace{2mm}9.4 & 52.2 & 49.0\\

\midrule
Oat-Zero & Math & Rule & 45.8 & \textbf{38.8} & 53.3  &  71.5 & 80.8   & 52.1 & \textbf{29.8} & 52.4 & 53.2\\
PRIME & Math & Rule & 39.5 & 32.1 & 47.7 & 54.5 & 76.4  & 45.5 & 20.4 & 43.4 & 45.2\\
SimpleRL-Zoo & Math & Rule & 46.9 & 38.4 & 51.1 & 70.3 & 77.1  & 51.0 & 26.5 & 51.7 & 51.6 \\
\midrule
TTRL & Base & Rule & 51.1 & 34.1 & 48.8 & 68.0 & \textbf{82.1} & 52.8 & 15.8 & 50.5 & 50.4 \\
SimpleRL-Zoo & Base & Rule & 54.1 & 36.2 & 49.5 & 70.7 & 76.3 &  49.2 & 14.8 & 52.6 & 50.1\\
RLVR & Base & Rule & 55.1 & 36.2 & 52.2 & 75.3 & 76.5 & 54.9 & 17.7 & 54.7 & 52.6 \\
General Reasoner & Base & Model & 55.4 & 37.4 & 52.1 & 74.5 & 77.0 & 51.7 & 16.0 & 54.8 & 52.0 \\
VeriFree & Base & \textcolor{teal}{\ding{55}} & 53.8 & 36.7 & 47.6 & 72.5 & 73.5 & 49.0 & 12.5 & 52.6 & 49.4 \\    
\rowcolor[HTML]{D7E8E8}
 \methodname~& Base & \textcolor{teal}{\ding{55}} &  \textbf{56.0} & 37.6 &  \textbf{55.4} & \textbf{75.5} & 78.0 & \textbf{56.5} & 16.3 & \textbf{56.1} & \textbf{53.6}\\
    \bottomrule
    \end{tabular}
    } % end resized box
    \caption{Overall performance on seven reasoning benchmarks. WebInst.: held-out evaluation set from WebInstruct. General: Average of MMLU-Pro, GPQA, TheoremQA and WebInst.}
    \label{tab:main_exp}
\end{table}

The main experimental results are reported in Table~\ref{tab:main_exp}, from which we observe that:
(1) \methodname~significantly improves general-domain reasoning performance. Without any external verifier, our method improves the average performance on four general-domain reasoning benchmarks by 24.9\% on Qwen2.5-7B.
(2) \methodname~exceeds the RLVR baseline on Qwen, Llama and Gemma. Specifically, we achieve larger general reasoning performance improvement over RLVR for 1.4, 3.9 and 1.4 average points on Gemma, Llama and Qwen respectively.
(3) \methodname~enhances mathematical reasoning capability on par with frameworks dedicated to math reasoning. Though we removed the mathematical category from the original WebInstrut dataset during training, we find the performance on multiple mathematical benchmarks is significantly improved and the score on Minerva surpasses Oat-Zero and SimpleRL-Zoo. 
(4) \methodname~exhibits even better performance compared with methods that require trained verifier models, surpassing General Reasoner, which uses a trained 1.5B-parameter verifier model to judge each sampled response, by 1.6 on average across all seven reasoning benchmarks. 
(5) \methodname~achieves a significant performance advantage compared with concurrent verifier-free methods, with improvement of 7.6 points on TheoremQA and 7.5 points on Minerva over VeriFree~\citep{zhou2025reinforcing}.

\subsection{Probability-based Reward Analysis}
\label{sec:pr_analysis}

\begin{wrapfigure}{r}{0.5\linewidth}
\centering
\vspace*{-4mm}
\includegraphics[width=0.5\textwidth]{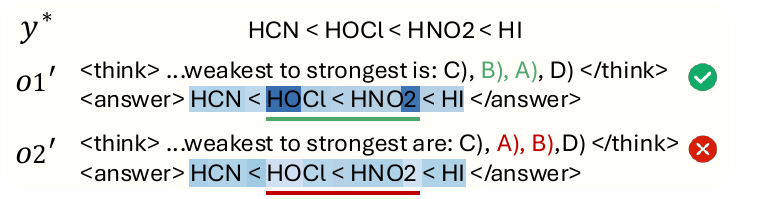}
\vspace{-5mm}
    \captionof{figure}{Token-level probability visualization, where deeper colors represent higher values. The underlined part highlights that probabilities precisely reflect that response sequence $o2$ incorrectly place option B after A, resulting lower scores at the corresponding positions in the reference answer. The question is shown in Figure~\ref{fig:teaser}.}
    \label{fig:pr_case}
\vspace*{-5mm}
\end{wrapfigure}

We first illustrate a token-level probability example in Figure~\ref{fig:pr_case}, where response sequence $o2$ receives a substantially lower score on the ``HO'' token, precisely reflecting the error made by response sequence  $o2$ (i.e., placing option A before option B). For quantitative analysis of the Probability-based Reward (PR) quality, we sample eight responses for each prompt from the WebInstruct~\citep{ma2025generalreasoneradvancingllmreasoning} and DeepScale~\citep{deepscaler2025} datasets.  To ensure a fair evaluation, we use the publicly released model from~\citep{hu2025openreasonerzeroopensourceapproach}. Human annotators then evaluate the correctness of each response. To maintain robustness and control labeling costs, we randomly keep 50 prompts from each dataset that contain both correct and incorrect responses.

\begin{wrapfigure}{r}{0.5\linewidth}
\centering
\vspace*{-5mm}
\includegraphics[width=0.5\textwidth]{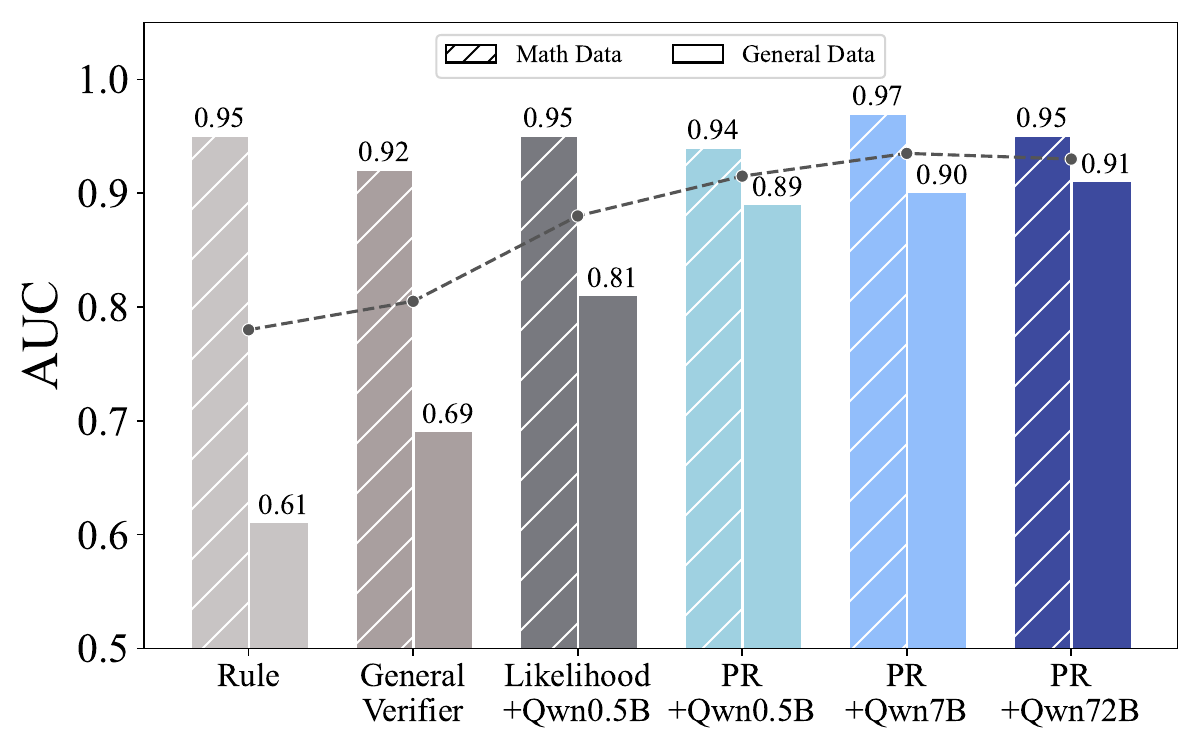}
\vspace{-5mm}
    \caption{Reward quality comparison. We report the AUC on both math data and general data, and highlight the average score with the dashed line. Qwn: Qwen2.5 models.}
    \label{fig:pr_auc}
\vspace*{-2mm}
\end{wrapfigure}

\textbf{PR discriminates correct responses better than the rule-based verifier on general data.}
To evaluate the ability of different reward to distinguish between correct and incorrect responses (i.e., assign higher rewards to correct responses), we rank responses for each prompt according to the respective rewards and compute the ROC-AUC~\citep{bradley1997use} metric using human annotations as ground truth. Higher AUC values indicate stronger discrimination capability. 
As shown in Figure~\ref{fig:pr_auc}, while the rule-based verifier achieves reasonable performance on mathematical prompts, it struggles on general-domain prompts, achieving an AUC of only 0.61. The primary flaw of the rule-based verifier in general domains is that it overlooks correct responses due to its limited capability of processing natural language complexity. We show an example in Figure~\ref{fig:teaser} to illustrate the phenomenon. In contrast, PR consistently delivers high-quality rewards across both mathematical and general domains.

\textbf{PR outperforms verifier models across both mathematical and general domains.} While the General-Verifier achieves improvement over rule-based reward on general data~(0.61$\rightarrow$0.69), its performance declines on mathematical prompts (0.95$\rightarrow$0.92) as shown in Figure~\ref{fig:pr_auc}. We attribute this limitation to the finetuning-based paradigm, which requires extensive task-specific data and struggles to generalize across domains. In contrast, our proposed PR achieves improvements of at least 2\% on mathematical data and 20\% on general-domain data compared with the verifier model. Upon analyzing the General-Verifier's judgments, we find that its main errors stem from limited comprehension of complex responses and challenges in output parsing. 
By leveraging the intrinsic capabilities of LLMs, PR directly produces high-quality reward scores in a single forward pass, also eliminating the need for any text post-processing.

\textbf{PR is effective with even small-scale models.} We compare the quality of PR using models of varying sizes. As shown in Figure~\ref{fig:pr_auc}, even the smallest Qwen2.5-0.5B outperforms the specifically trained General-Verifier on both mathematical and general data. While increasing the model size further improves the performance on general-domain data, gains on mathematical data are marginal due to the already high absolute scores.

\begin{wraptable}{r}{0.4\textwidth}
\centering
\vspace*{-4mm}
\resizebox{\linewidth}{!}{
\begin{tabular}{lc|cc}
\toprule
\multirow{2}{*}{Data}  & \multirow{2}{*}{Verifier} &  TheoremQA & Minerva\\
& & Avg@2 & Avg@2  \\
\midrule
DAPO & Rule  & 50.3  & 50.6 \\
\midrule
\multirow{2}{*}{WebInstruct}  & Rule   &  52.2 & 54.9  \\
  & \textcolor{teal}{\ding{55}}  & \textbf{55.4} & \textbf{56.5} \\
\bottomrule
\end{tabular}
}
\caption{Effect of different RLVR training data and reward mechanisms. 
% All results are trained from Qwen2.5 7B.
}
\label{tab:data_comparison}
\vspace*{-4mm}
\end{wraptable}

\textbf{PR is robust over entropy and length distribution.} We also analyze the robustness of PR by analyzing the correlation between PR values and factors, including length and decoding entropy of generated responses. For each prompt, we calculate the Spearman correlation coefficient and p-value. We observe that only 8\% prompts get a p-value smaller than 0.05, and the average coefficient is -0.060 for length and 0.059 for entropy.
These results indicate that the probability reward values show negligible correlation with both entropy and length. This indicates that our proposed reward serves as a robust reward mechanism.

\textbf{PR is essential for utilizing general-domain data.}
We compare the performance of models trained exclusively on mathematical prompts~\cite{yu2025dapo} versus those trained on general-domain prompts, as shown in Table~\ref{tab:data_comparison}. The results demonstrate that general-domain data enhances the performance on both benchmarks (+1.9 on TheoremQA, +4.3 on Minerva). However, general-domain data also includes additional challenges for rule-based verifiers. Consequently, directly adopting existing rule-based verifiers gives obvious diminished performance.

\subsection{Ablation Study}

To investigate the contribution of different design choices in \methodname, we perform an ablation study.

\textbf{Effect of per-token probability as reward.} We compare our per-token probability-based reward with naive sequence likelihood as the reward signal. In the calculation of likelihood, low-probability tokens can dramatically affect the final reward. 
For instance, probabilities of $1\text{e}^{-4}$ versus $1\text{e}^{-5}$ can lead to a tenfold difference in reward, despite their small absolute difference. This issue becomes more pronounced for longer reference answers, which are more likely to contain at least one low-probability token.
\citep{zhou2025reinforcing} addresses this instability by filtering out prompts whose reference answers exceed seven tokens. However, this also significantly limits the data diversity. In contrast, using the mean per-token probability is much more robust and yields better performance, as shown in Table~\ref{tab:ablation}. 
We also compare the reward quality of the likelihood reward and our proposed PR in Figure~\ref{fig:pr_auc}, where PR consistently achieves better results on both domains.

\textbf{Effect of reward debiasing and standard deviation filtering.} We compare our final debiased reward $\hat{r}$ with directly using the reward in Eq~\ref{eq:prob_score}. Results in Table~\ref{tab:ablation} shows that the performance on both benchmarks is worse with original reward, demonstrating the effectiveness of the debiasing operation.  To quantify the effectiveness of the standard deviation filtering approach, we also train a model without any filtering mechanism. The results in Table~\ref{tab:ablation} show that the filtering strategy is important for the final performance of models by removing prompts that do not get diverse responses.

\begin{minipage}{0.48\textwidth}
\centering
\resizebox{\linewidth}{!}{
\begin{tabular}{lcc}
\toprule
\multirow{1}{*}{Method}  &  TheoremQA & Minerva\\
\midrule
\methodname~ & \textbf{55.4} & \textbf{56.5} \\
\hspace{2mm} w/o debiasing & \hspace{2mm}\deltavalue{52.7}{2.7}{1.7}& \hspace{2mm}\deltavalue{54.1}{2.4}{1.7}\\
\hspace{2mm} w/o std-filtering & \hspace{2mm}\deltavalue{52.5}{2.9}{1.7}& \hspace{2mm}\deltavalue{55.1}{1.4}{1.7}\\
\hspace{2mm} w/o token prob. & \hspace{3mm}\deltavalue{33.5}{21.9}{1.7}  & \hspace{3mm}\deltavalue{34.2}{22.3}{1.7} \\ % 
% \hspace{1mm} + RLAIF-V BoN &  7B  & \hspace{1mm}LLaVA-NeXT\hspace{1mm} & \deltavalue{6.8}{3.7}{1.7} & \deltavalue{3.8}{1.4}{1.7} & \deltavalue{39.7}{4.8}{0} & \deltavalue{3.07}{0.1}{0} & \deltavalue{{28.1}}{4.2}{0} &  N/A & N/A  & N/A & \deltavalue{55.7}{8.5}{0} & \deltavalue{24.4}{1.9}{0} \\
\bottomrule
\end{tabular}
}
\captionof{table}{Ablation experimental results. Token prob.: token probability average. Avg@2 results are reported.}
\label{tab:ablation}
\end{minipage}%
\hfill
\begin{minipage}{0.48\textwidth}
\centering
\vspace*{-3mm}
\resizebox{\linewidth}{!}{
\begin{tabular}{lcc}
\toprule
\multirow{1}{*}{Reward}  &  TheoremQA & Minerva\\
\midrule
Rule-based  & 44.8 & \textbf{50.0} \\
Rule-based + PR & \textbf{48.8}  & 49.0 \\
\bottomrule
\end{tabular}
}
\captionof{table}{Experimental results of different rewards on mathematical data. Avg@2 results are reported. We combine rule-based reward and PR by summarizing advantages.}
\label{tab:combine_pr_vr}
\end{minipage}

\subsection{\methodname~on verifiable domains}

We study the effectiveness of \methodname~on domains where verifiers are already available. In this section, we use the mathematical training data of PRIME~\citep{cui2025entropy} as a representative mathematical RLVR dataset. Though rule-verifiers already give a reliable correctness label on mathematical data, we observe that such a binary correctness label lacks fine-grained discrimination capability on different responses sharing the same correctness. For example, given reference answer ``200'' for a question, ``199'' is generally better than ``1''. We argue that such fine-grained discrimination can be helpful for the model to get a more comprehensive understanding of the qualities of sampled responses and thus improve its performance. We combine the rule-based verifier scores and our proposed PR to train the model and report results in Table~\ref{tab:combine_pr_vr}. Results show that our proposed probability reward can also improve the utilization of data from verifiable domains like mathematics.

\subsection{Robustness Analysis}

\begin{figure}
    \centering
    \vspace*{-5mm}
    \includegraphics[width=\linewidth]{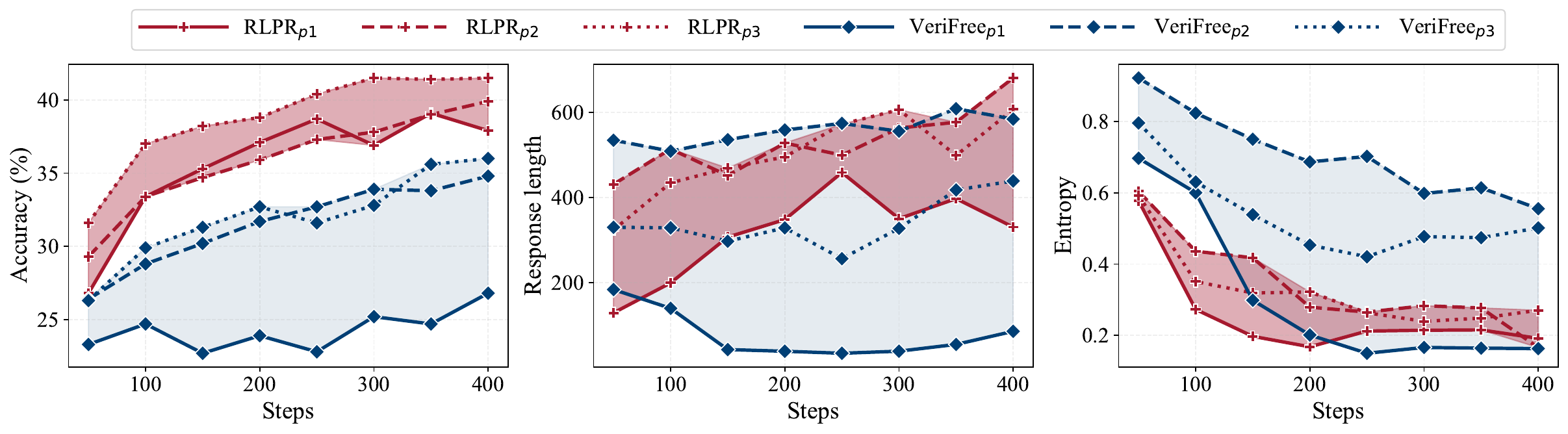}
    \vspace{-5mm}
    \caption{Robustness across different training prompt templates. \textcolor[HTML]{a6192e}\methodname~yields consistently higher performance compared with \textcolor[HTML]{003f75}{VeriFree}. Left: average performance on seven benchmarks. Middle: response length. Right: response entropy during training. }
    \label{fig:robustness}
    \vspace*{-5mm}
\end{figure}

Compared with rule-based rewards, the distribution of our proposed probability-based reward (PR) may be influenced by variations in training prompt templates. To evaluate the robustness of \methodname~with different templates, we consider three prompt settings: $p_1$ from VeriFree~\cite{zhou2025reinforcing}, $p_2$ used in DeepSeek-R1~\cite{deepseekai2025deepseekr1incentivizingreasoningcapability} and $p_3$ which moves the format requirement to user prompt. To reduce training costs, we switch the base model to Qwen2.5-3B, decrease the batch size to 128, and apply a single update per training step. For fair comparison, we adopt the origin dataset from VeriFree for this experiment. Figure~\ref{fig:robustness} presents the comparison of performances, response length, and entropy across different training steps. We observe that \methodname~maintains consistent performance regardless of prompt choice, while VeriFree exhibits high sensitivity, with a notable performance drop of by 8.0 at step-400 when using $p_1$. Furthermore, the response length of \methodname~under all prompts converges to a similar level, and the entropy remains within a reasonable range with no signs of entropy collapse~\cite{cui2025entropy}.

\section{Related Works}

\textbf{Reinforcement Learning with Verifiable Rewards}. Reinforcement learning from binary verifiable rewards~\citep{cui2025process,yu2025dapo,luo2025deepscaler,qwq32b,deepseekai2025deepseekr1incentivizingreasoningcapability} recently demonstrates strong reasoning capabilities on math and code tasks, and has emerged as a common practice. 
These practices utilize verifiers such as Math-Verify~\citep{mathverify}, SandboxFusion~\citep{bytedanceseedfoundationcodeteam2025fullstackbenchevaluatingllms}, and custom implemented ones~\citep{cui2025process}, which effectively judge the correctness of model rollouts and forgo the need for preference annotations. 
However, this paradigm is restricted to domains where robust verifiers are available. 
Moreover, existing implementations of verifiers show inconsistencies~\citep{skywork-or1-2025} since the complexity for rule-based verifiers to handle edge cases is nontrivial. 
In this work, we propose to extend RLVR practices to domains without robust verifiers. 

\textbf{Reasoning in General Domains}
Previous research explores reasoning in general domains, a vital part of which is how to obtain reliable reward signals. One line of work is generative reward models~\citep{mahan2024generative}, where another generative model judges the quality of rollouts. This concept has been extended to the implementation of verifiers based on a generative model~\citep{ma2025generalreasoneradvancingllmreasoning,liu2025x} and enhancements of the judge model itself as a reasoner~\citep{chen2025rm}. 
In this work, we demonstrate that reinforcement learning for general-domain reasoning can rely on the decoding probability of the reference answer as a reward signal. 
Concurrent to our work, ~\citep{zhou2025reinforcing} utilizes policy likelihood for reference answer as rewards, while limited to short answers less than $7$ tokens and requires a auxiliary fine-tuning-based objective. 
Instead, we observe the robustness of per-token probability as a reward signal and extend RLVR to general domains without length constraints.

\textbf{Self-Reward Optimization}
Unsupervised reinforcement learning on language models using the policy model itself as a reward has recently emerged as an embarrassingly effective approach~\citep{zuo2025ttrltesttimereinforcementlearning,zhao2025learning}. 
The common idea behind the practice of self-reward is raising the probability of consistent answers~\citep{zuo2025ttrltesttimereinforcementlearning}, intuitively from the observation that concentrating on the majority brings free improvements~\citep{wang2022self}. 
Recent literature~\citep{agarwal2025unreasonable} shows that entropy minimization, which naively degrades generation diversity, is a sugar for reasoning tasks, though restricted to certain model families. 
However, such practice might be problematic for restricting exploration~\citep{cui2025entropy,hochlehnert2025sober,yu2025dapo}. 
In contrast to self-rewarding methods that remove diversity to exploit existing reasoning ability, our approach builds the reward based on the reference answer, yielding reasoning performance with healthy token entropy from the clip-high trick~\citep{yu2025dapo}.

\section{Conclusion}

RLVR shows the power of scaling test-time computation for addressing complex problems and sheds valuable light on paths to AGI. In this work, we present \methodname~, a novel framework that extends this paradigm to broader general domains. Comprehensive experimental results on Gemma, Llama and Qwen show that our method achieves significant improvement on both general and mathematical reasoning tasks without using external verifiers. We propose a novel probability reward (PR) and reward debiasing strategy to enhance its quality further. By replacing rule-based reward with PR, we eliminate the need for external verifiers and achieve better performance than using naive likelihood as a reward or using verifier models.  Moreover, we propose a simple standard deviation filtering strategy that stabilizes training by removing samples with low reward standard deviation.  In the future, we will explore more domains, including multimodal understanding and scaling \methodname~to larger models.

\bibliography{iclr2025_conference}
\bibliographystyle{iclr2025_conference}

\newpage
\appendix
\section{Appendix}

\subsection{Experimental Details}
Our experiments are conducted on Qwen2.5-7B~\citep{qwen2.5} if not additionally specified. 
Following most RLVR practices, we forgo the supervised fine-tuning process and directly post-train on the base model, and use GRPO algorithm by default. 
We change the prompt template during training and validation time in our main experiments to control the response structure to have extractable thoughts and answers. 
The prompt template is shown in Table~\ref{tab:train_prompt}.

\begin{table*}[h]
\centering
\begin{minipage}{0.99\linewidth}\vspace{0mm}    \centering

\begin{tcolorbox}[colframe=black!75!white, colback=white, coltitle=white, title=\methodname training prompt, fonttitle=\bfseries]
\small
\begin{verbatim}
<|im_start|>system
A conversation between User and Assistant. The user asks a question,
and the Assistant solves it. The assistant first thinks about the 
reasoning process in the mind and then provides the user with the 
answer. The reasoning process and answer are enclosed within <think>
</think> and <answer> </answer> tags, respectively, i.e., <think>
reasoning process here </think> <answer> answer here </answer>.
<|im_end|>
<|im_start|>user
{{question}}<|im_end|>
<|im_start|>assistant

\end{verbatim}
\end{tcolorbox}
\caption{We adopt the training prompt of R1~\citep{deepseekai2025deepseekr1incentivizingreasoningcapability} for \methodname.}

\label{tab:train_prompt}
\end{minipage}
\end{table*}

\subsubsection{Parameter Settings}
Each experiment is trained on 32 NVIDIA A100 GPUS. 
We use a 1e-3 entropy penalty coefficient and no KL penalty. 
The learning rate for the policy model is 5e-7. 

\subsubsection{Training Logs}
\begin{figure}[h]
    \centering
  \begin{subfigure}{0.32\linewidth}
    \centering
    \includegraphics[width=\linewidth]{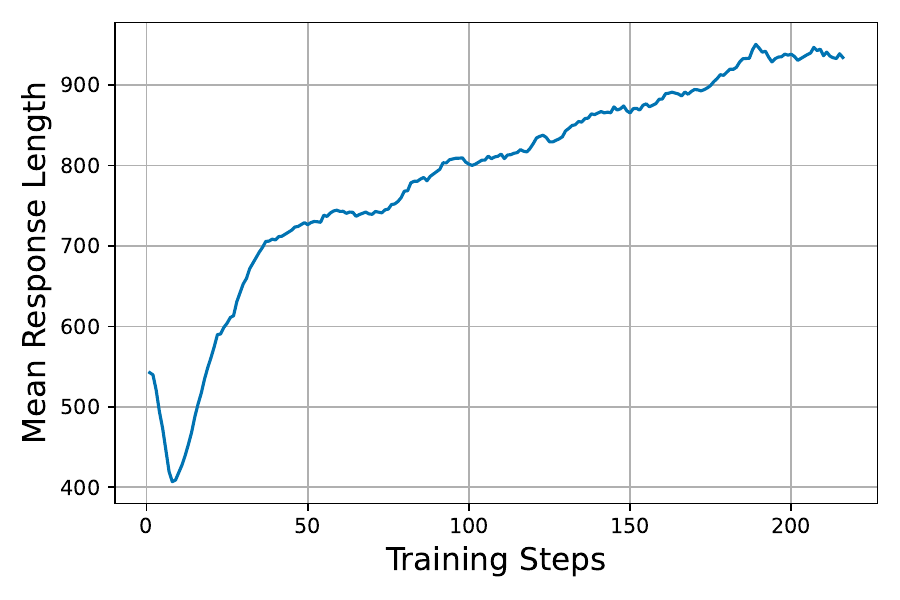}
    \caption{Mean response length}
    \label{fig:dynamics_length}
  \end{subfigure}
  \hfill
    \begin{subfigure}{0.32\linewidth}
    \centering
    \includegraphics[width=\linewidth]{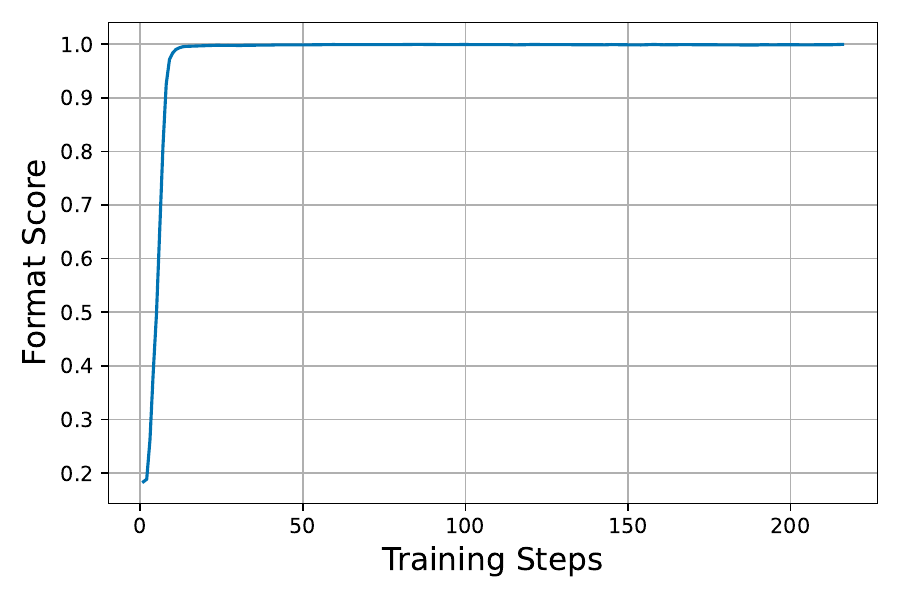}
    \caption{Format reward}
    \label{fig:dynamics_format}
  \end{subfigure}
  \hfill
    \begin{subfigure}{0.32\linewidth}
    \centering
    \includegraphics[width=\linewidth]{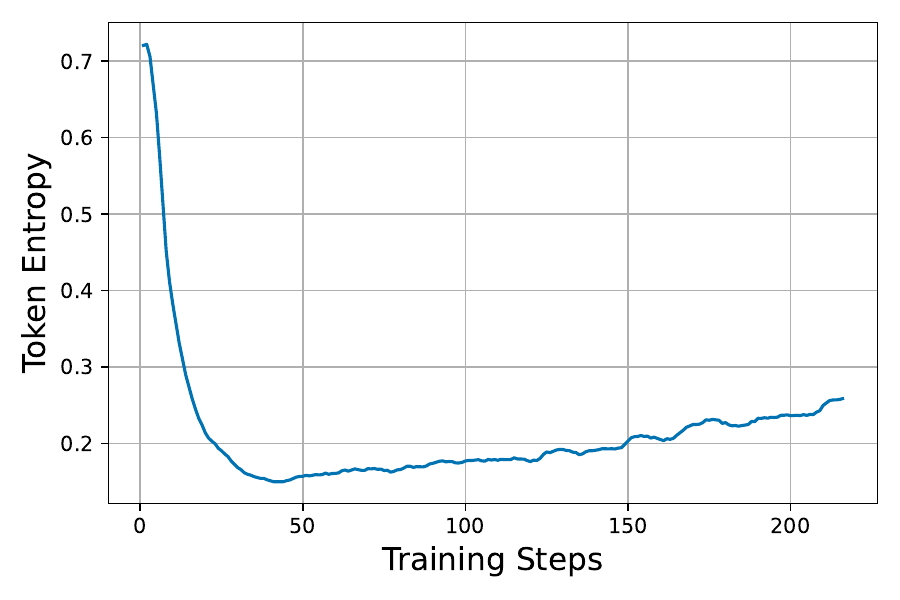}
    \caption{Token entropy}
    \label{fig:dynamics_entropy}
  \end{subfigure}

    \caption{Training dynamics of \methodname on Qwen2.5-7B}
    \label{fig:dynamics}
\end{figure}

We monitor key training metrics of our methods in~\ref{fig:dynamics}. 
During training, the response length~(Fig. \ref{fig:dynamics_length}) steadily increases, allowing more profound reasoning behaviors and no sign of degeneration. 
In Fig. \ref{fig:dynamics_format}, the policy model quickly learns to follow the response structure. 
Moreover, as shown in Fig. \ref{fig:dynamics_entropy}, our training entropy exhibits neither collapses as a result of the clip-high trick, nor abrupt increases. This ensures the balance between exploration and exploitation.

\begin{figure}[t]
    \centering
    \includegraphics[width=0.46\textwidth]{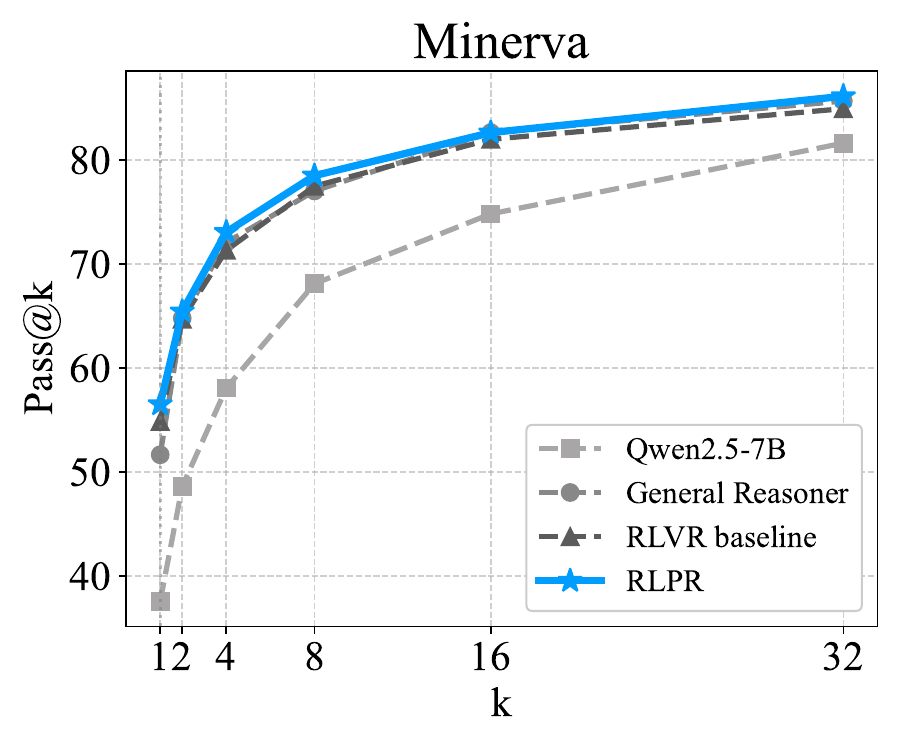}\hfill
    \includegraphics[width=0.46\textwidth]{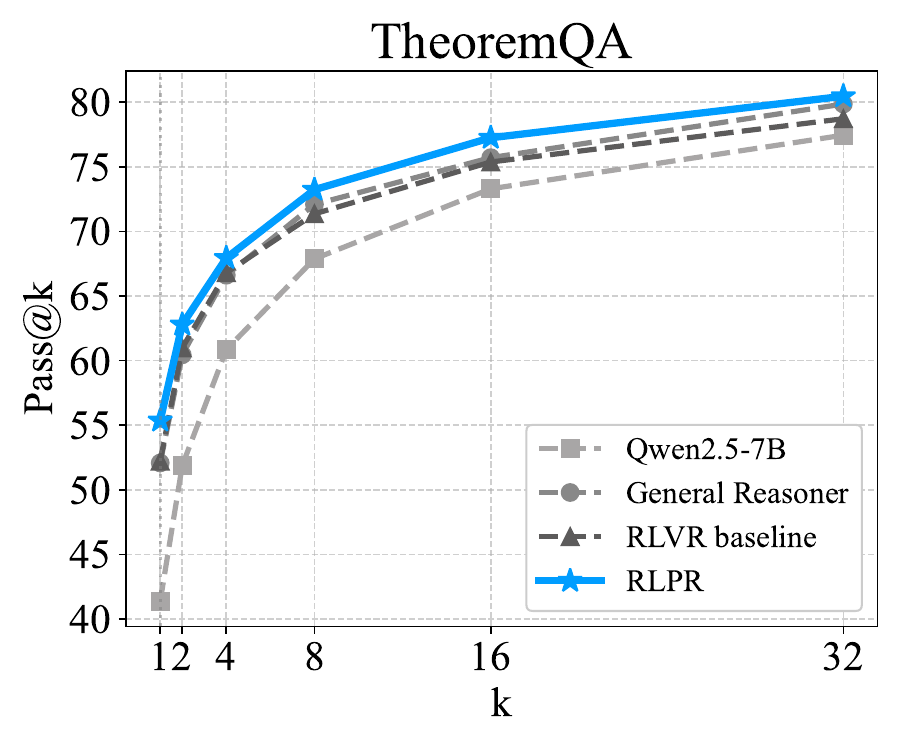}\hfill
    \caption{Pass@k curves for \methodname and baselines. }
    \label{fig:passatk}
\end{figure}

\subsection{Pass@k Evaluation}
To further test the impact on the potential reasoning boundary of our method, we provide pass@k results on various tasks in Fig.~\ref{fig:passatk}.
Compared to standard RLVR and General Reasoner, \methodname shows comparable or better pass@k accuracy, 
indicating that our method is not trading reasoning potential for pass@1 improvements.

\subsection{Training Data}

We adopt WebInstruct~\citep{ma2025generalreasoneradvancingllmreasoning} as our training dataset, excluding  math-related prompts to focus on general-domain reasoning. To ensure the quality and difficulty of training samples, we apply a multi-stage filtering strategy: First, we remove history-related questions and those targeting elementary or middle school levels to avoid commonsense or overly simple content. Finally, leveraging GPT-4.1-mini’s reasoning scores (1–4, see Table~\ref{tab:gpt41_reasoning_score}), we retain only highly challenging samples (score $\geq$ 3). This process reduces the dataset from 231,833 to 77,687 samples, yielding a focused and high-quality corpus for complex non-mathematical reasoning.

\begin{table*}[t]
\centering
\begin{minipage}{0.99\linewidth}\vspace{0mm}    \centering

\begin{tcolorbox}[colframe=black!75!white, colback=white, coltitle=white, title=Prompt for GPT-4.1 to assess reasoning complexity, fonttitle=\bfseries]
\small

\# Description

\vspace{1mm}
You are asked to evaluate the reasoning level requirement of problems. Problems are scored from 1 to 4, with higher scores indicating greater reasoning demands. You should make your decision based on the following detail instructions.

\vspace{2mm}
\#\# 1 Point: No reasoning requirements.

\vspace{1mm}
Problems requiring direct recall of specific facts and commonsense knowledge.

Examples:\\
- What is Fermat's Last Theorem. (Requires only recalling facts)\\
- What are the five quantitative forecasting models? (Requires only recalling facts)\\
- What is the capital of China? (Requires only recalling commonsense knowledge)\\
- When was Mark Twain born? (Requires only recalling commonsense knowledge)\\

\vspace{2mm}
\#\# 2 Points: No reasoning skill requirements.

\vspace{1mm}
Problems that do not require reasoning skills. Either because (1) it is too simply and reasoning skills don't help, or (2) it is too hard to clearly rank different answers since the question is too open-ended and reasoning skills also do not help.\\

Examples:\\
- Solve x + 1 = 10, what is the value of x. (Too simple)\\
- What would you do if you have four legs. (Too open-ended to determine response quality)\\

\#\# 3 Points: Moderate level reasoning skills and knowledge are enough.

\vspace{1mm}
Problems requiring moderate level reasoning skills and knowledge. Such as problems that are mostly likely to be solved by any random undergraduate student regardless of their majors.\\

Examples:\\
- Solve a quadratic equation: x² - 5x + 6 = 0.\\
- Find all solutions to $[\sqrt{x} + 2 \sqrt{x^2 + 7x} + \sqrt{x + 7} = 35 - 2x]$. Enter all the solutions, separated by commas.\\
- Summarize the main causes of World War I. (Requires recalling and organizing established historical factors).\\
- Describe the importance of empathy in storytelling to someone unfamiliar with the concept, using no more than 4 sentences, and ensure all text is in lowercase. Include a quote from a famous author at the end.\\

\#\# 4 Points: Long-time analysis and deep understanding of relevant knowledge are required.

\vspace{1mm}
Problems requiring long-time to analyze and solve, and depend on deep understanding of relevant knowledge. Such as designing a complex system, developing a comprehensive strategy or providing detail and easy-to-understand solution for realworld problems.\\

Examples:\\
- Design a scalable and secure REST API for a large e-commerce platform, considering microservices architecture, data consistency, fault tolerance, and evolving business needs.\\
- Develop a comprehensive urban planning strategy for sustainable development in a rapidly growing city, integrating environmental, social, economic, and infrastructural considerations.\\
- Conduct a thorough root cause analysis for a major systemic failure (e.g., a financial crisis or a large-scale environmental disaster) and propose multi-level preventative and corrective policy measures.\\
- The polynomial $P(x)=(1+x+x^2+\ldots+x^{17})^2-x^{17}$ has 34 complex zeros of the form $z_k=r_k\left[\cos(2\pi\alpha_k)+i\sin(2\pi\alpha_k)\right]$, $k=1,2,3,\ldots,34$, with $0<\alpha_1\le\alpha_2\le\alpha_3\le\dots\le\alpha_{34}<1$ and $r_k>0$. Find $\alpha_1+\alpha_2+\alpha_3+\alpha_4+\alpha_5.$ \\

Please score the following question:
Q: \texttt{\{question\}}

You should first explain your reasoning briefly, then give the final score in following format:

Reasoning score: [1-4]

\end{tcolorbox}
\caption{Prompt for GPT-4.1 to assess reasoning complexity.}

\label{tab:gpt41_reasoning_score}
\end{minipage}
\end{table*}

\subsection{Implementation Details}\label{sec:impl_details_appendix}

\begin{table}
\centering
    \resizebox{\linewidth}{!}{
    \setlength\tabcolsep{2.2pt}
    \begin{tabular}{llcccc}
    \toprule
    \textbf{Experiment name} & \textbf{Table / Figure} & \textbf{Batch Size} & \textbf{Update per Step} & \textbf{Clip Threshold} & \textbf{$\beta$} \\
    \midrule
    \multirow{4}{*}{Main experiment} &  Figures~\ref{fig:performance_viper}, ~\ref{fig:dynamics}, ~\ref{fig:passatk} & 768 & 4 & (0.8, 1.27) & 0.5 \\
    & Qwen in Table~\ref{tab:main_exp} & 768 & 4 & (0.8, 1.27) & 0.5 \\
    & Llama in Table~\ref{tab:main_exp} & 256 & 4 & (0.8, 1.27) & 0.9 \\
    & Gemma in Table~\ref{tab:main_exp} & 256 & 4 & (0.8, 1.27) & 1.0 \\
    \midrule
    RLPR vs. RLPR & Table~\ref{tab:data_comparison} & 768 & 4 & (0.8, 1.27) & 0.5 \\
    Ablation study  & Table~\ref{tab:ablation} & 768 & 4 & (0.8, 1.27) & 0.5 \\
    \midrule
    RLPR on verifiable domains & Table~\ref{tab:combine_pr_vr} & 128 & 2 &  (0.8, 1.20) & - \\
    \midrule
    Robustness analysis & Figure~\ref{fig:robustness} & 128 & 1 & (0.8, 1.27) & 0.5 \\
    \bottomrule
    \end{tabular}
}
    \caption{Implementation setup for each experiment. Default settings align with Sections~\ref{sec:exp_setup} and~\ref{sec:impl_details_appendix}.}
    \label{tab:implementation_details}
\end{table}

This section provides additional implementation details to supplement Section~\ref{sec:exp_setup}. The policy model generates 8 responses per question, using a learning rate of 1e-6. We remove the KL divergence term by setting the KL coefficient to 0. Detailed configurations are presented in Table~\ref{tab:implementation_details}, where the number of policy updates per step and the value of $\beta$ are empirically determined to be optimal for their respective scenarios.  

RLVR baselines are trained under the same setting with corresponding \methodname~results, except using rule-based verifiers and accuracy filtering. For RLVR training on Llama and Gemma, we find accuracy filtering can remove over 90\% training prompts and thus significantly increase the training cost and find small batch size causes entropy blow up. So we do not apply accuracy filtering for these two experiments and conduct only one update for each batch to stabilize training.

\end{document}